\documentclass[10pt,twocolumn,letterpaper]{article}

\usepackage{wacv}
\usepackage{times}
\usepackage{epsfig}
\usepackage{graphicx}
\usepackage{amsmath}
\usepackage{amssymb}
\usepackage{times}
\usepackage{epsfig}
\usepackage{graphicx}
\usepackage{amsmath}
\usepackage{amssymb}
\usepackage{adjustbox}
\usepackage{tabularx}
\usepackage[caption=false]{subfig}
\usepackage{capt-of}

\usepackage{color}
\usepackage[dvipsnames]{xcolor}
\usepackage{enumitem}

\usepackage{xcolor} 
\usepackage{soul} 
\usepackage{xspace}
\usepackage{amsmath}

\soulregister{\ref}{7}
\soulregister{\cite}{7}
\soulregister{\subsection}{1}
\soulregister{\section}{1}
\soulregister{\label}{1}

\newcommand{\mylossRna}{$\mathcal{L}_{RNA}$ }

\newcommand{\cmark}{\ding{51}}%
\newcommand{\xmark}{\ding{55}}%

\usepackage{booktabs}
\usepackage{multirow}
\usepackage{colortbl}
\definecolor{Gray}{gray}{0.9}

\makeatletter
\@namedef{ver@everyshi.sty}{}
\makeatother
\usepackage{tabularx}

\usepackage{tikz}
\usepackage{graphicx}
\usepackage{pgfplots}
\usepackage{adjustbox}
\usepackage[accsupp]{axessibility} 

\usepackage{xcolor-material}
\usepackage{multirow}
\usepackage{colortbl}

\usepackage{etoolbox}
\usepackage{amssymb}
\usepackage{pifont}

\DeclareMathOperator{\EX}{\mathbb{E}}

\def\wacvPaperID{624} 

\wacvfinalcopy 

\ifwacvfinal
\fi

\ifwacvfinal
\usepackage[breaklinks=true,bookmarks=false,colorlinks]{hyperref}
\else
\usepackage[pagebackref=true,breaklinks=true,colorlinks,bookmarks=false]{hyperref}
\fi

\begin{document}

\title{Domain Generalization through  Audio-Visual  Relative Norm Alignment \\ in First Person Action Recognition}

\author{
Mirco Planamente\thanks{The authors equally contributed to this work.  }\textsuperscript{ }\textsuperscript{          ,1,2,3}
 \quad
Chiara Plizzari\footnotemark[1]\textsuperscript{ }\textsuperscript{    ,1} \quad

Emanuele Alberti\textsuperscript{1} \quad
Barbara Caputo\textsuperscript{1,2,3}
\and \textsuperscript{1} Politecnico di Torino

\and \textsuperscript{2} Istituto Italiano di Tecnologia

\and \textsuperscript{3}  CINI Consortium
\and {\{\tt\small mirco.planamente, chiara.plizzari, emanuele.alberti,  barbara.caputo\}@polito.it}

}




\maketitle

\ifwacvfinal
\thispagestyle{empty}
\fi


\begin{abstract}

First person action recognition is becoming an increasingly researched area thanks to the rising popularity of wearable cameras. This is bringing to light cross-domain issues that are yet to be addressed in this context. Indeed, the information extracted from learned representations suffers from an intrinsic ``environmental bias". This strongly affects the ability  to generalize to unseen scenarios, limiting the application of current methods to real settings where labeled data are not available during training. 
In this work, we introduce the first domain generalization approach for egocentric activity recognition, by proposing a new audio-visual loss, called Relative Norm Alignment loss. It re-balances the contributions from the two modalities during training, over different domains, by aligning their feature norm representations.
Our approach leads to strong results in domain generalization on both EPIC-Kitchens-55 and EPIC-Kitchens-100, as demonstrated by extensive experiments, and can be extended to work also on domain adaptation settings with competitive results.

\end{abstract}

\section{Introduction}

\begin{figure}[t]
    \centering
    \includegraphics[width=0.97\linewidth]{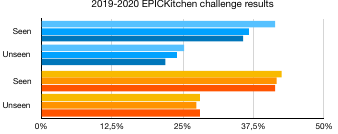}
    \caption{Top-3 results of the \textcolor{RoyalBlue}{2019} \cite{ek19report} and \textcolor{orange}{2020} \cite{ek2020} EK challenges, when testing on ``Seen" and ``Unseen" kitchens.}
    \label{fig:ek-challenge}
\end{figure}

First Person Action Recognition is rapidly attracting the interest of the research community~\cite{sudhakaran2018attention,sudhakaran2019lsta,furnari2020rolling,Kazakos_2019_ICCV,ghadiyaram2019large,Wu_2019_CVPR,fathi2011learning,rodin2021predicting, li2018eye}, both for the significant challenges it presents and for its central role in real-world egocentric vision applications, from wearable sport cameras to human-robot interaction or human assistance. 
The recent release of the EPIC-Kitchens large-scale dataset \cite{damen2018scaling}, as well as the competitions that accompanied it, has encouraged research into more efficient architectures. 
In egocentric vision, the recording equipment is worn by the observer and it moves around with her. Hence, there is a far higher degree of changes in illumination, viewpoint and environment compared to a fixed third person camera.
Despite the numerous publications in the field, egocentric action recognition still has one major flaw that remains unsolved, known as ``environmental bias” \cite{torralba2011unbiased}.
This problem arises from the network's heavy reliance on the environment in which the activities are recorded, which inhibits the network's ability to recognize actions when they are conducted in unfamiliar (unseen) surroundings.
To give an intuition of its impact, we show in Figure \ref{fig:ek-challenge} the relative drop in model performance from the seen to unseen test set of the top-3 methods of the 2019 and 2020 EPIC-Kitchens challenges.
\begin{figure}[t]
\centering

    \includegraphics[width=\linewidth]{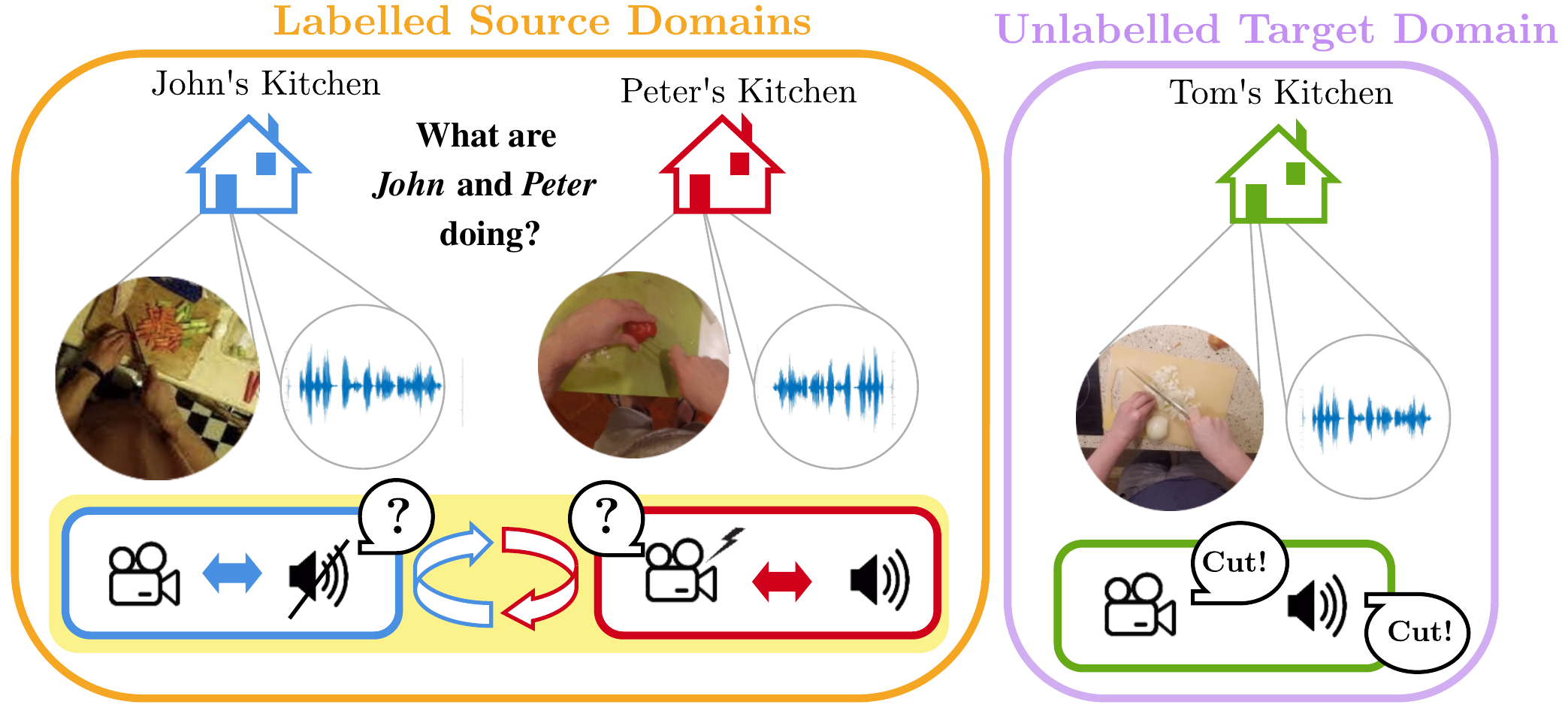}
    \caption{
    Due to their intrinsic different nature, audio and visual information suffer from the domain shift in different ways. However, the network tends to ``privilege" one modality over the other. 
    We re-balance the contribution of the two modalities while training, allowing the network to learn ``equally" from each.} 

\label{fig:teaser}
\vspace{-0.2cm}

\end{figure}
In general, this problem is referred to in the literature as \textit{domain shift}, meaning that a model trained on a source labelled dataset cannot generalize well on an unseen dataset, called target.
Recently, \cite{Munro_2020_CVPR} addressed this issue by reducing the problem to an unsupervised domain adaptation (UDA) setting, where an unlabeled set of 
samples from the target is available during training. However, the UDA scenario is not always realistic, because the target domain might not be known a priori or because accessing target data at training time might be costly (or plainly impossible).

In this paper, we argue that the true challenge is to learn a representation able to generalize to any unseen domain, regardless of the possibility to access target data at training time. This is known as domain generalization (DG) setting. 
Inspired by the idea of exploiting the multi-modal nature of videos \cite{Munro_2020_CVPR,Kazakos_2019_ICCV}, we make use of multi-sensory information to deal with the challenging nature of the setting.
Although the optical flow modality is the most widely utilized \cite{Munro_2020_CVPR, wang2016temporal,sudhakaran2019lsta, furnari2020rolling}, it requires a high computational cost, limiting its use in online applications. Furthermore, it may not be ideal in a wearable context where battery and processing power are restricted and must be preserved.
The audio signal has the compelling advantage of being natively provided by most wearable devices \cite{kumar2018wearable}, and thus it does not require any extra processing.
Egocentric videos come with rich sound information, due to the strong hand-object interactions and the closeness of the sensors to the sound source, and audio is thus suitable for first person action recognition  \cite{Kazakos_2019_ICCV,cartas2019seeing,kazakos2021slow}. 
Moreover, the ``environmental bias” impacts auditory information as well, but in a different way than it affects visual information. In fact, audio and video originate from distinct sources, i.e., camera and microphone. We believe that the complementarity of the two can help to attenuate the domain shift they both suffer.
For instance, the action ``cut" presents several audio-visual differences across domains: cutting boards will differ in their visual and auditory imprints (i.e., wooden cutting board \textit{vs} plastic one), various kinds of food items might be cut, and so forth (Figure~\ref{fig:teaser}).


Despite multiple modalities could potentially provide additional information, the CNNs' capability to effectively extract useful knowledge from them is somehow restricted  \cite{gradient-blending,alamri2019audio,goyal2017making,poliak2018hypothesis,weston2011wsabie}. 
The origin of this difficulty, in our opinion, is due to one modality being ``privileged" over the other during training.
Motivated by these findings, 
we propose the Relative Norm Alignment loss, a simple yet effective loss whose goal is to re-balance the mean feature norms of the two modalities during training, allowing the network to fully exploit joint training, especially in cross-domain scenarios. 
To  summarize,  our  contributions  are  the  following: 
\begin{itemize}
    \item we bring to light the “unbalance” problem arising from training multi-modal networks, which causes the network to “privilege” one modality over the other during training, limiting its generalization ability;
    \item we propose a new cross-modal audio-visual loss, the Relative Norm Alignment (RNA) loss, that progressively aligns the relative feature norms of the two modalities from various source data, resulting in domain-invariant audio-visual features;
    \item  we present a new benchmark for 
    multi-source domain generalization in first person videos and extensively validate our method on both DG and UDA scenarios. 
\end{itemize}
\section{Related Works}

\textbf{First Person Action Recognition.} The main architectures utilized in this context, which are generally inherited from third-person literature, divide 
into two categories: methods based on 2D convolution \cite{10.5555/2968826.2968890, wang2016temporal, ma2016going, lin2019tsm, zhou2018temporal, Kazakos_2019_ICCV, cartas2019seeing, sudhakaran2020gate} and method based on 3D ones \cite{carreira2017quo, singh2016first, Wu_2019_CVPR, kapidis2019multitask, tran2015learning, feichtenhofer2019slowfast, 10.5555/2968826.2968890, Munro_2020_CVPR}. LSTM or its variations \cite{Sudhakaran_2017_ICCV, sudhakaran2018attention, sudhakaran2019lsta, furnari2020rolling, planamente2021self} commonly followed the first group to better encode temporal information.
The most popular technique is the multi-modal approach \cite{carreira2017quo, Munro_2020_CVPR, wang2016temporal,sudhakaran2019lsta, furnari2020rolling}, especially in EPIC-Kitchens competitions \cite{damen2020rescaling, damen2018scaling}. Indeed, RGB data is frequently combined with motion data, such as optical flow.
However, although optical flow has proven to be a strong asset for the action recognition task, it is computationally expensive. As shown in \cite{Crasto_2019_CVPR}, the use of optical flow limits the application of several methods in online scenarios, pushing the community either towards single-stream architectures~\cite{zhao2019dance,Crasto_2019_CVPR,lee2018motion,sun2018optical,planamente2021self}, or to investigate alternative modalities, e.g., audio information \cite{kazakos2021slow}. 
Although the audio modality has been proven to be robust in egocentric scenarios by \cite{Kazakos_2019_ICCV,cartas2019seeing, kazakos2021slow}, this work is the first to exploit it, jointly with its visual counterpart, in a cross-domain context.

\textbf{Audio-Visual Learning.}
Many representation learning methods use self-supervised approaches to learn cross-modal representations that can be transferred well to a series of downstream tasks.
Standard tasks are Audio-Visual Correspondence~\cite{look_listen_learn, cooperative_torresani,objects_that_sound}, or Audio-Visual Synchronization \cite{chung2016out,afourasself,multisensory_owens,korbar2018cooperative}, 
which was shown to be useful for sound-source localization \cite{objects_that_sound,afourasself,Zhao_2018_ECCV,senocak2018learning,tian2018audio}, active speaker detection \cite{chung2016out,afourasself} and multi-speaker source separation \cite{multisensory_owens,afourasself}. 
Other audio-visual approaches have been recently proposed \cite{listen_to_look,morgado2020learning,tian2020unified,Morgado_2021_CVPR,morgado2021audio,alwassel2019self,korbar2018co} which exploit the natural correlation between audio and visual signals.
Audio has also attracted attention in egocentric action recognition \cite{Kazakos_2019_ICCV,cartas2019seeing} and has been used in combination with other modalities \cite{gradient-blending}. 
However, none of these techniques has been intended to cope with cross-domain scenarios, whereas this paper demonstrates the audio modality's ability to generalize to unseen domains when combined with RGB.




\textbf{Unsupervised Domain Adaptation (UDA).}
We can divide UDA approaches into \textit{discrepancy-based} methods, which explicitly minimize a distance metric among source and target distributions \cite{da-afnxu2019larger, da-mcdsaito2018maximum,da-mmdlong2015learning}, 
and \textit{adversarial-based} methods~\cite{da-adv-deng2019cluster, da-adv-tang2020discriminative}, often leveraging a gradient reversal layer (GRL)~\cite{grl-pmlr-v37-ganin15}. 
Other works exploit batch normalization layers to normalize source and target statistics \cite{ada-bn, DBLP:conf/iclr/LiWS0H17, da-bnchang2019domain}.
The approaches described above have been designed for standard image classification tasks. Other works analyzed UDA for video tasks, including action detection \cite{agarwal2020unsupervised}, segmentation \cite{chen2020action} and classification \cite{videoda-chen2019temporal,Munro_2020_CVPR,videoda-choi2020unsupervised,videoda-Jamal2018DeepDA,pan2020adversarial,Song_2021_CVPR}. 
Recently \cite{Munro_2020_CVPR} proposed an UDA method for first person action recognition, called MM-SADA, consisting of a combination of existing DA methods trained in a multi-stage fashion. 


\textbf{Domain Generalization (DG).} 
Previous approaches in DG are mostly designed for image data \cite{carlucci2019domain,volpi2018generalizing,li2018domain,dou2019domain,li2018deep,bucci2020selfsupervised} and are divided in \textit{feature-based} and \textit{data-based} methods. The former focus on extracting invariant information which are shared across-domains~\cite{li2018domain,li2018deep}, while the latter exploit data-augmentation strategies to augment source data with adversarial samples to get closer to the target~\cite{volpi2018generalizing}. Interestingly, using a self-supervised pretext task is an efficient solution for the extraction of a more robust data representation \cite{carlucci2019domain,bucci2020selfsupervised}. We are not aware of previous works on first or third person DG. 
Among unpublished works, we found only one \textit{arXiv} paper~\cite{videodg-yao2019adversarial} in third person action recognition, designed for single modality. Under this setting, first person action recognition models, and action recognition networks in general, degenerate in performance due to the strong divergence between source and target distributions. 

Our work stands in this DG framework, and proposes a feature-level solution to this problem in first person action recognition by leveraging  audio-visual correlations. 

\section{ Proposed Method } \label{Method}
In this work, we bring to light that the discrepancy between the \textit{\textbf{mean feature norms}} of audio and visual modalities negatively affects the training process of multi-modal networks, leading to sub-optimal performance. Indeed, it causes the modality with greater feature norm to be ``privileged" by the network 
during training, while ``penalizing" the other. We refer to this problem with the term \textit{\textbf{``norm unbalance"}}. The intuitions and motivations behind this problem, as well as our proposed solution to address it, are presented below.
 
\subsection{Intuition and Motivation}\label{sec:intuition}

A common strategy in the literature to solve the first-person action recognition task is to use a multi-modal approach \cite{Munro_2020_CVPR,Kazakos_2019_ICCV,lin2019tsm,cartas2019seeing,kazakos2021slow,wang2016temporal}.
Despite the wealth of information of multi-modal networks w.r.t. the uni-modal ones, their performance gains are limited and not always guaranteed \cite{gradient-blending,alamri2019audio,goyal2017making,poliak2018hypothesis,weston2011wsabie}. 
Authors of \cite{gradient-blending}
 attributed this problem to overfitting, and addressed it by re-weighting the loss value of each stream through different hyperparameters. This technique, however, necessitates a precise estimating step, which is dependent on the task and the dataset.
In this paper, we approach the above described multi-modal issue from a different perspective, by considering an audio-visual framework.

\begin{figure}[t]
    \includegraphics[width=0.97\linewidth]{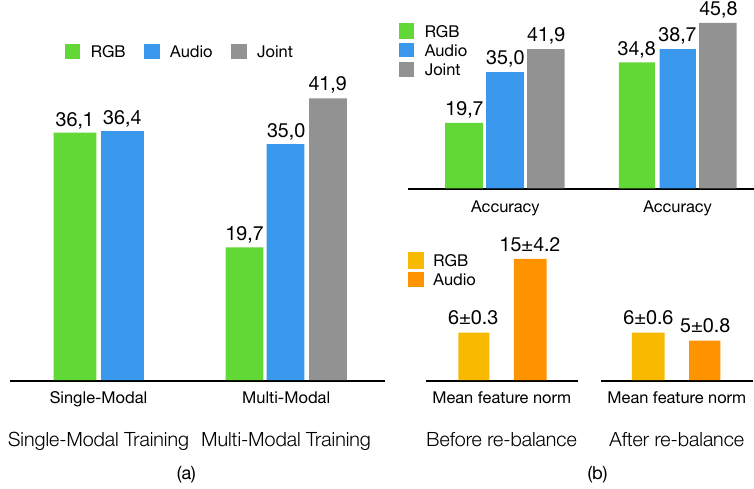}
    \caption{By jointly training, and testing on separate streams, the RGB performance drop (left). “unbalance" at feature-norm level which, when mitigated, leads to better performance (right).} 
    \label{fig:norm_ranges}
    \vspace{-0.3cm}
\end{figure} 

\textbf{Norm \textit{unbalance}.} We hypothesize that during training there is an ``unbalance" between the two modalities that prevents the network from learning ``equally" from the two. 
This hypothesis is also supported by the fact that the hyperparameters discovered in~\cite{gradient-blending} differ significantly depending on the modality.
To empirically confirm this intuition, we performed a simple experiment, which is shown in Figure \ref{fig:norm_ranges}-a.
Both modalities perform equally well at test time when RGB and audio streams are trained independently.
However, when trained together and tested separately, the RGB accuracy decreases compared to the audio accuracy. This proves that the optimization of the RGB stream was negatively affected by multi-modal training.
We also wondered whether this concept, i.e., the unbalance that occurs between modalities during the training phase, could be extended to a multi-source context.
Is it possible that one source has a greater influence on the other, negatively affecting the final model?
Based on the above considerations, we searched for a function that captures the amount of information contained in the final embedding of each modality, possibly justifying the existence of this unbalance.

\textbf{The \textit{mean feature norms}.} Several works highlighted the existence of a strong correlation between the mean feature norms and the amount of ``valuable" information for classification \cite{zheng2018ring,wang2017normface,ranjan2017l2}. In particular, the cross-entropy loss has been shown to promote well-separated features with a high norm value in \cite{wang2017normface}. 
Moreover, the work of \cite{ye2018rethinking} is based on the Smaller-Norm-Less-Informative assumption, which implies that a modality representation with a smaller norm is less informative during inference.
All of the above results suggest that the $L2$-norm of the features gives an indication of their information content, and thus can be used as a metric to measure the unbalance between the two training modalities. 
By studying the behaviour of the feature norms, we found that, on the training set, the mean feature norms of audio samples ($\simeq$ 32) was larger than that of RGB ones ($\simeq$10). This unbalance is also reflected on the test set (Figure \ref{fig:norm_ranges}-b, left), with the modality with the smaller norm being the one whose performance are negatively affected.

Motivated by these results, we propose a simple but effective loss whose goal is to re-balance the mean feature norms during training across multiple sources, so that the network can fully exploit joint training, especially in cross-domain scenarios. In fact, when re-balancing the norms, 
the performance of both modalities increase (Figure \ref{fig:norm_ranges}-b, right).
Note that the concept of the smaller norm being less informative is used to argue that the network's preference for the audio modality is only due to its higher norm (over the RGB one), but this does not imply that RGB is less informative for the task; indeed, the range of norms after re-balancing is closer to the original RGB norm.



\subsection{Domain Generalization} 


We assume to have different source domains $\{\mathcal{S}_1,...,\mathcal{S}_k\}$, where each $\mathcal{S}={\{(x_{s,i},y_{s,i})\}}^{N_s}_{i=1}$ is composed of $N_s$ source samples with label space $Y_s$ known, and a target domain $\mathcal{T}={\{x_{t,i}\}}^{N_t}_{i=1}$ of $N_t$ target samples whose label space $Y_t$ is unknown. 
The objective is to train a model able to predict an action of the target domain without having access to target data at training time, thus exploiting the knowledge from multiple source domains to improve generalization.
The main assumptions are that the distributions of all the domains are different, i.e., $\mathcal{D}_{s,k} \neq \mathcal{D}_t$ $\land$ $\mathcal{D}_{s,k} \neq \mathcal{D}_{s,j}$, with $k \neq j$, $k,j=1,...,k$, and that the label space is shared, $\mathcal{Y}_{s} = \mathcal{Y}_{t}$. 


\subsection{Framework}\label{RNA-NET}
\begin{figure*}[t!]
    \centering
    \includegraphics[width=1\linewidth]{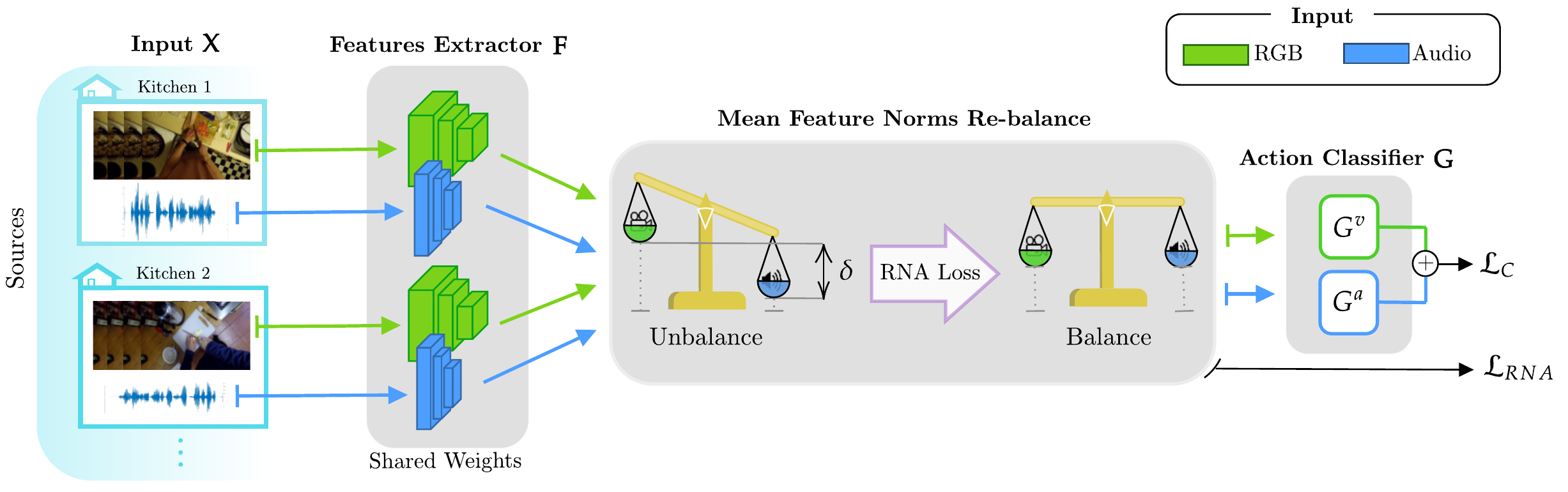}
    
    \caption{\textbf{RNA-Net.} Labeled \textcolor{OliveGreen}{source visual} $x^v_{s,i}$ and \textcolor{RoyalBlue}{source audio} $x^a_{s,i}$ inputs are fed to the respective feature extractors $F^v$ and $F^a$. Our loss $\mathcal{L}_\textit{RNA}$ operates at the feature-level by balancing the relative feature norms of the two modalities. The network is trained with a standard cross-entropy loss $\mathcal{L}_{c}$ jointly with $\mathcal{L}_\textit{RNA}$. At inference time, multi-modal target data is used for classification.
    }
    
    
    \label{fig:architecture}
\end{figure*}

Let us consider an input $x=(x^v_i,x^a_i)$, where we denote with $v$ and $a$ the visual and audio modality respectively, and with $i$ the $i$-th sample. 
As shown in Figure \ref{fig:architecture}, each input modality ($x^v_i,x^a_i$) is fed to a separate feature extractor, $F^v$ and $F^a$ respectively. 
The resulting features $f^v=F^v(x^v_i)$ and $f^a=F^a(x^a_i)$ are then passed to the separate classifiers $G^v$ and $G^a$, whose outputs correspond to distinct score predictions (one for each modality). They are then combined with a \textit{late fusion} approach to obtain a final prediction (see Section \ref{sec:experimental} for more details). 
The whole architecture, which we call RNA-Net, is trained by minimizing the total loss, defined as
\begin{equation}\label{eq:total_loss}
    \mathcal{L}= \mathcal{L}_C + \lambda \mathcal{L}_{RNA},
\end{equation}       
where the $ \mathcal{L}_C $ is the standard \textit{cross-entropy loss} and $\lambda$ indicates the weight given to the proposed cross-modal loss, called Relative Norm Alignment loss ($\mathcal{L}_{RNA}$). 
Technical details of $\mathcal{L}_{RNA}$ are defined in the next section.





\subsection{Relative Norm Alignment Loss} \label{sec:rna_loss}
\begin{figure}[t]
    \includegraphics[width=1\linewidth]{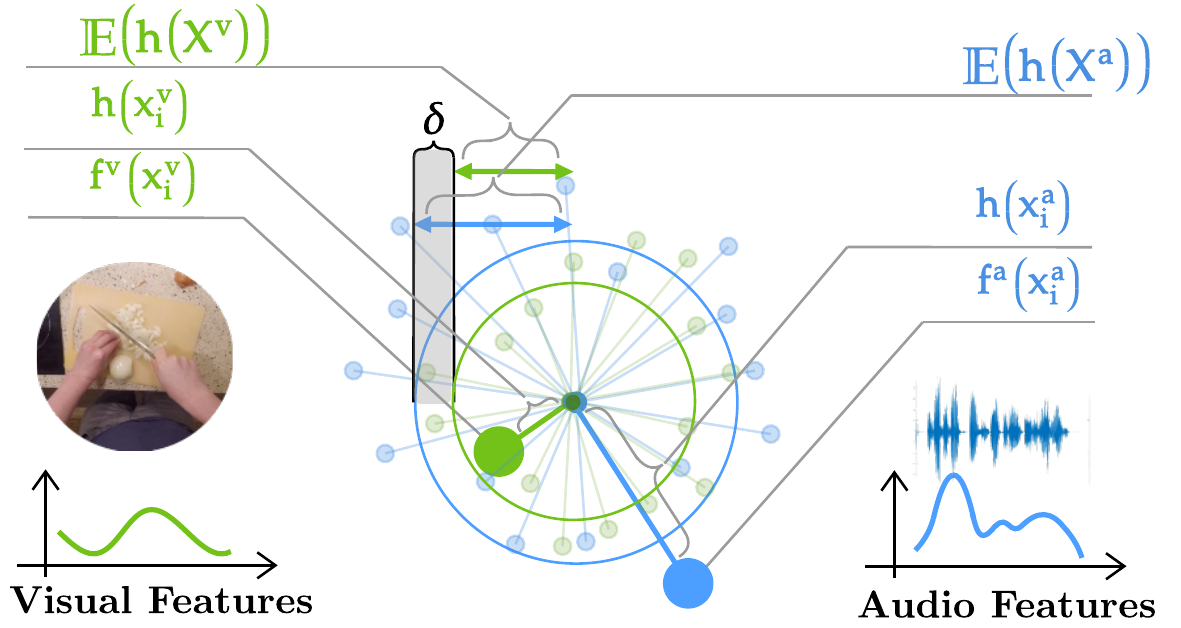}

    \caption{
    The norm $h(x^v_i)$ of the $i$-th \textcolor{OliveGreen}{visual} sample (left) and $h(x^a_i)$ of the $i$-th \textcolor{RoyalBlue}{audio} sample (right) are represented, by means of segments of different lengths. The radius of the two circumferences represents the mean feature norm of the two modalities, and $\delta$ their discrepancy. 
    By minimizing $\delta$, audio and visual feature norms are induced to be the same.}
    \label{fig:norm}
\end{figure}

\textbf{Definition.} The main idea behind our loss is the concept of \textit{mean-feature-norm distance}.
We denote with $h(x^m_i)=({\lVert{ \cdot }\rVert}_2 \circ f^m)(x^m_i)$ the $L_2$-norm of the features $f^m$ of the $m$-th modality,
and compute the \textit{mean-feature-norm distance} ($\delta$) between the two modality norms $f^v$ and $f^a$ as
\begin{equation}\label{eq:RNA}
    \delta(h(x^v_i),h(x^a_i))=\lvert{\EX[h(X^v)]-\EX[h(X^a)]\rvert}
\end{equation} 
where $\EX[h(X^m)]$ corresponds to the mean features norm for each modality.
Figure \ref{fig:norm} illustrates the norm $h(x^v_i)$ of the $i$-th visual sample and $h(x^a_i)$ of the $i$-th audio sample, by means of segments of different lengths arranged in a radial pattern. The mean feature norm of the $k$-{th} modality is represented by the radius of the two circumferences, and $\delta$ is represented as their difference. The objective is to minimize the $\delta$ distance by means of a new loss function, which aims to align the mean feature norms of the two modalities. {In other words, we restrict the features of both modalities to lie on a hypersphere of a fixed radius.  }

We propose a Relative Norm Alignment loss, defined as
\begin{equation}\label{formula:rna}
    \mathcal{L}_{RNA}=\left(\frac{\EX[h(X^v)]}{\EX[h(X^a)]} - 1\right)^2,
\end{equation}
where $\EX[h(X^m)]=\frac{1}{N}\sum_{x^m_i \in \mathcal{X}^m}h(x^m_i)$ for the $m$-th modality and $N$ denotes the number of samples of the set $\mathcal{X}^m=\{x^m_1,...,x^m_N\}$. This dividend/divisor structure is introduced to encourage a relative adjustment between the norm of the two modalities, inducing an \textit{optimal equilibrium} between the two embeddings. {Furthermore, the square of the difference pushes the network to take larger steps when the ratio of the two modality norms is too far from one, resulting in faster convergence. }



Conceptually, aligning the two modality norms corresponds to imposing a ``hard" constraint, aligning them to a constant value $k$. We refer to this as \textit{Hard Norm Alignment} (HNA), and we formulate the corresponding $\mathcal{L}_\textit{HNA}$ loss as 
\begin{equation}\label{eq:hna}
    \mathcal{L}_\textit{HNA}=\sum_m\left(\EX[h(X^m)] - k\right)^2,
\end{equation}
where $k$ is the same for all $m$ modalities. 
Nevertheless, as shown in Section \ref{exp}, our formulation of $\mathcal{L}_\textit{RNA}$ helps the convergence of the two distributions' norms without requiring this additional $k$ hyper-parameter. 
Designing the loss as a \textit{subtraction} ($\mathcal{L}_{RNA}^{sub}$) between the two norms by directly minimizing $\delta^2$ (Equation \ref{eq:RNA}) (see Supplementary) is a more straightforward solution and a valid alternative.
However, the rationale for this design choice is that a substantial discrepancy between the value of $k$ and $\EX[h(X^m)]$, as well as $\EX[h(X^v)]$ and $\EX[h(X^a)]$, would reflect in a high 
loss value, thus requiring an accurate tuning of the weights and consequently increasing the network sensitivity to loss weights~\cite{kendall2018multi}. 
Indeed, this dividend/divisor structure ensures that loss to be in the 
range $(0,1]$, starting from the modality with higher norm as dividend.




\textbf{\textit{Learn} to re-balance.}\label{properties}
The final objective of RNA loss is to \textit{learn} how to leverage audio-visual norm correlation \textit{at feature level} for a general and effective classification model. It is precisely because the network learns to solve this task that we obtain generalization benefits, rather than avoiding the norm unbalance through input level normalization or pre-processing. Note that introducing a normalization at \textit{input level} could be potentially not suitable for pre-trained models. Moreover, it would not be feasible in DG, where the access to target data is not available during training, and thus not only there is no information on the target distribution, but each domain also requires a distinct normalization.

Additionally, the rationale behind \textit{learning} to re-balance rather than using typical projection methods to normalize features~\cite{ranjan2017l2} is two-fold. First, forcing the network to normalize the features using model weights themselves mitigates the ``norm unbalance" problem also during inference, as the network has the chance to learn to work in the normalized feature space during training. Secondly, explicit normalization operators, e.g., batch normalization, impose the scaled normal distribution individually for each element in the feature. However, this does not ensure that overall mean feature norm of the two modalities to be the same.

\subsection{Extension to Unsupervised Domain Adaptation}\label{uda}
Thanks to the unsupervised nature of $\mathcal{L}_{RNA}$, our network can be easily extended to the UDA scenario. 
Under this setting, both labelled source data from a single source domain $\mathcal{S}=(\mathcal{S}^v,\mathcal{S}^a)$, and unlabelled target data $\mathcal{T}=(\mathcal{T}^v$, $\mathcal{T}^a)$ are available during training. 
We denote with $x_{s,i}=(x^v_{s,i},x^a_{s,i})$ and $x_{t,i}=(x^v_{t,i},x^a_{t,i})$ the $i$-th source and target samples respectively.  Both $x^m_{s,i}$ and $x^m_{t,i}$ are fed to the feature extractor $F^m$ of the $m$-th specific modality, shared between source and target, obtaining respectively the features $f_s=(f^v_s,f^a_s)$ and $f_t=(f^v_t,f^a_t)$. In order to consider the contribution of both source and target data during training, we redefine our \mylossRna under the UDA setting as 
\begin{equation}\label{eq:loss_s_t}
    \mathcal{L}_{RNA}=\mathcal{L}^s_{RNA}+\mathcal{L}^t_{RNA} 
\end{equation}
where $\mathcal{L}^s_{RNA}$ and $\mathcal{L}^t_{RNA}$ correspond to the RNA formulation in Equation \ref{eq:RNA}, applied to source and target data respectively. Both the contributions are added in Equation \ref{eq:total_loss}.



\section{Experiments}\label{exp}

In this section, we first introduce the dataset used and the experimental setup (Section \ref{sec:exp_setting}), then we present the experimental results (Section \ref{results}). We compare RNA-Net against existing multi-modal approaches, and both DG and UDA methods. 
We complete the section with some ablation studies and qualitative results.

\label{sec:experimental}
\subsection{Experimental Setting}\label{experimental_setting}\label{sec:exp_setting}

\noindent
\textbf{Dataset.} 
We use  the EPIC-Kitchens-55 dataset \cite{damen2018scaling} and we adopt the same experimental protocol of \cite{Munro_2020_CVPR}, where the three kitchens with 
the largest amount of labeled samples are handpicked from the 32 available. We refer to them here as D1, D2, and D3 respectively. 

\noindent
\textbf{Input.}
Regarding the RGB input, a set of 16 frames, referred to as \textit{segment}, is randomly sampled during training, while at test time 5 equidistant segments spanning across all clips are fed to the network. At training time, we apply random crops, scale jitters and horizontal flips for data augmentation, while at test time only center crops are applied. 
Regarding aural information, we follow
\cite{Kazakos_2019_ICCV} and convert the audio track into a 256 $\times$ 256 matrix representing the log-spectrogram of the signal. The audio clip is first extracted from the video, sampled at 24kHz and then the Short-Time Fourier Transform (STFT) is calculated of a window length of 10ms, hop size of 5ms and 256 frequency bands.


\noindent
\textbf{Implementation Details.} Our network is composed of two streams, one for each modality $m$, with distinct feature extractor $F^{m}$ and classifier $G^{m}$. The RGB stream uses I3D \cite{carreira2017quo}
as done in
\cite{Munro_2020_CVPR}. 
The audio feature extractor uses the BN-Inception model \cite{bn-inception} pretrained on ImageNet \cite{imageNet}, which proved to be a reliable backbone for the processing of audio spectrograms~\cite{Kazakos_2019_ICCV}. Each $F^{m}$ produces a 1024-dimensional representation $f_{m}$ which is fed to the classifier $G^{m}$, consisting in a fully-connected layer that outputs the score logits. Then, the two modalities are fused by summing the outputs and the cross entropy loss is used to train the network. 
We train RNA-Net for $9k$ iterations using the SGD optimizer. The learning rate for RGB is set to $1e-3$ and reduced to $2e-4$ at step $3k$, while for audio, the learning rate is set to $1e-3$ and decremented by a factor of $10$ at steps $\{1000,2000,3000\}$. The batch size is set to $128$, and the weight $\lambda$ of $\mathcal{L}_{RNA}$ is set to $1$.

\subsection{Results}\label{results}


\newrobustcmd*{\mytriangle}[1]{\tikz{\filldraw[draw=#1,fill=#1] (0,0) --
(0.2cm,0) -- (0.1cm,0.2cm);}}

\setlength\heavyrulewidth{0.31ex}

\begin{table*}[ht]

\begin{minipage}[t]{0.66\linewidth}
\centering
\begin{adjustbox}{width=\columnwidth}\centering

\begin{tabular}{lllll}

\toprule\noalign{\smallskip}
\multicolumn{5}{c}{\normalsize\textsc{Multi Source DG}} \\
\noalign{\smallskip}
\cline{1-5}
\noalign{\smallskip}
  & \multicolumn{1}{c}{D2, D3  $\rightarrow$ D1}  & D3, D1  $\rightarrow$ D2 & D1, D2  $\rightarrow$ D3 & Mean \\ 
 \noalign{\smallskip} \hline
 \noalign{\smallskip}
Deep All         &   43.19   & 39.35  &       51.47      & 44.67 \\ \hline
\noalign{\smallskip}
IBN-Net   \cite{instance_bn}  &  44.46   & 49.21 &     48.97 & 47.55   \\ 

MM-SADA (Only SS)   \cite{Munro_2020_CVPR}   &  39.79   & 52.73 &     51.87  & 48.13   \\ 

Gradient Blending   \cite{gradient-blending}   & 41.97  & 48.40 &     51.43 & 47.27   \\ 
\hline \noalign{\smallskip} 
TBN  \cite{Kazakos_2019_ICCV}    &  42.35  &  47.45 & 49.20    &  46.33  \\

Transformer    \cite{morgado2020learning}     &  42.78   & 47.38 &     51.79 & 47.32   \\

Cross-Modal Transformer    \cite{cheng2020look}       &  40.87   & 43.57 &     54.88  & 46.44   \\

SE  \cite{squeeze_and_excitation}     &  42.82  & 42.81 &    51.07 &  45.56  \\ 

Non-Local   \cite{non-local}   &  45.72   & 43.08 &     49.49 & 46.10   \\ 


 \hline
\noalign{\smallskip}

RNA-Net (Ours)  
&  45.65 \mytriangle{Green}\textcolor{Green}{$\mathbf{+2.46}$}   
&   51.64 \mytriangle{Green}\textcolor{Green}{$\mathbf{+12.32}$} 
&      55.88 \mytriangle{Green}\textcolor{Green}{$\mathbf{+4.41}$}
&  \textbf{51.06} \mytriangle{Green}\textcolor{Green}{$\mathbf{+6.4}$}\\ 
\bottomrule
\end{tabular}

\end{adjustbox} 
\vspace{+0.05cm}
\caption{Top-1 Accuracy ($\%$) of RNA-Net in Multi Source DG scenario. In \textcolor{Green}{\textbf{green}} we highlight improvement of RNA-Net w.r.t. the baseline Deep All.}
\label{tab:mm-1}
\end{minipage}
\hspace{2em}
\begin{minipage}[t]{0.26\linewidth}

\begin{adjustbox}{width=\linewidth}

\begin{tabular}{lc}
\toprule\noalign{\smallskip}
\multicolumn{2}{c}{\normalsize\textsc{UDA }} \\
\noalign{\smallskip}
\cline{1-2}
\noalign{\smallskip}
Method & Mean \\
 \noalign{\smallskip} \hline
 \noalign{\smallskip}
\noalign{\smallskip}
 Source-Only & 41.87 \\       
\noalign{\smallskip} \hline
 \noalign{\smallskip}

MM-SADA (Only SS) \cite{Munro_2020_CVPR}    &  46.44  \\ 
\noalign{\smallskip}

GRL \cite{grl-pmlr-v37-ganin15}     &   43.67     \\ 
\noalign{\smallskip}

MMD \cite{da-mmdlong2015learning}     &  44.46        \\ 
\noalign{\smallskip}

AdaBN \cite{ada-bn}   &  41.92        \\ 
\noalign{\smallskip}

RNA-Net (Ours)   & \textbf{47.71}  \\ \hline
\noalign{\smallskip}

MM-SADA (SS+GRL) \cite{Munro_2020_CVPR}       & 47.75        \\ 
\noalign{\smallskip}

RNA-Net+GRL (Ours)      & \textbf{48.30 } \\ 

\bottomrule

\end{tabular}
\end{adjustbox}
\vspace{+0.05cm}

\caption{Top-1 Accuracy ($\%$) of RNA-Net in UDA context.}
\label{tab:mm-2}
\end{minipage}


\end{table*}
\noindent

\begin{table*}[ht]
\begin{minipage}{\linewidth}

\begin{adjustbox}{width=1\columnwidth, margin=0ex 1ex 0ex 0ex}
\begin{tabular}{lccccccc|cccc}
\toprule



 & \multicolumn{2}{c}{Target: D1} & \multicolumn{2}{c}{Target: D2} & \multicolumn{2}{c}{Target: D3}   &  \\ \hline\noalign{\smallskip}

 & \multicolumn{1}{c}{D2  $\rightarrow$ D1  } & D3 $\rightarrow$ D1    & D1 $\rightarrow$  D2   & D3  $\rightarrow$ D2   & D1  $\rightarrow$ D3    & D2  $\rightarrow$ D3   & Mean  & D2, D3  $\rightarrow$ D1 & D3, D1  $\rightarrow$ D2 &  D1, D2  $\rightarrow$ D3 & Mean\\
\hline
\noalign{\smallskip}
Baseline (RGB Only) &    	34.76 &		33.03 & 34.15 &	41.08 &	35.03 &38.79	& 36.14  & 39.72 &  33.59   &37.91    & 37.07\\
Baseline (Audio Only) &   	29.17 & 34.00 & 31.90&		42.93  &	36.49 &	44.00 &	36.42  & 41.57 & 39.21 &   47.19 &    42.66 \\
\hline 
\noalign{\smallskip}
Baseline &   35.27&	\underline{40.26}&	 39.03&	\underline{49.98} &	39.17	& \underline{47.52}  &	41.87                     &   43.19   & 39.35 &        51.47       & 44.67\\
\noalign{\smallskip}

\hline
\noalign{\smallskip}

RNA-Net  &        41.76 &    \underline{42.20}  &  45.01                    &        \underline{51.98}  &  44.62   &   \underline{48.90}       & \textbf{45.75}                    &   {45.65}    &   51.64 &          55.88        &  \textbf{51.06} \\ 

\noalign{\smallskip}

\hline
 \noalign{\smallskip}

\rowcolor{Gray}
Best Single-Source  &  \xmark  &   Best D1     &  \xmark  & Best D2     &   \xmark    &       Best D3          &       & 40.26 &  49.98   & {47.52}       & 45.92  \\ 
\rowcolor{Gray}
Best Single-Source + RNA   &  \xmark  &   Best D1     &  \xmark  & Best D2     &   \xmark    &       Best D3      &            & 42.20 & 51.98   & 48.90        &47.69    \\ 

\bottomrule
\end{tabular}

\end{adjustbox}
\end{minipage}
\caption{ Top-1 Accuracy (\%) of RNA-Net w.r.t. uni-modal, multi-modal baseline and the \textit{Best Single-Source} both w/o and w RNA loss. \textbf{Bold}: highest mean result, \underline{underline} the best Single-Source case. } 
\label{tab:ablations}
\end{table*}

\textbf{DG Results. } Table \ref{tab:mm-1} illustrates the results of RNA-Net under the multi-source DG setting. We 
compare it to the \textit{Deep All} approach, namely
the backbone architecture when no other domain adaptive strategies are exploited and all the source domains are fed to the network. Indeed, this is 
the baseline in all image-based DG methods \cite{bucci2020selfsupervised}. 
We adapted a well-established image-based domain generalization approach, namely IBN-Net \cite{instance_bn}, and the multi-modal self-supervised task proposed in MM-SADA \cite{Munro_2020_CVPR} to evaluate the effectiveness of RNA-Net in the DG scenario. Indeed, 
training the network to solve a self-supervised task jointly with the classification has been demonstrated to be helpful in generalizing across domains \cite{bucci2020selfsupervised}. Finally, we compare RNA-Net against Gradient Blending (GB) \cite{gradient-blending}. 
The results in Table \ref{tab:mm-1} show that 
RNA-Net outperforms all the above mentioned methods by a significant margin. 
\begin{figure}[t]
    \includegraphics[width=\linewidth]{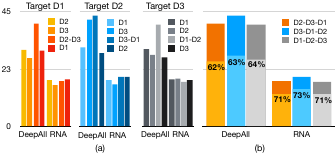}
    
    \caption{Mean feature norms of target when training on different source domains. The norm unbalance reflects also between different source domains, and RNA mitigates it (left). The percentage of the norm of the most relevant features over the total norm increases when minimizing RNA loss (right). 
}
    \label{fig:norm_unbalance}
\end{figure}

\textit{What are the benefits of RNA loss in generalizing? }
The RNA loss prevents the network from 
overspecializing across all domains from which it receives input.
As it can be seen from Figure \ref{fig:norm_unbalance}-a, the total norm (RGB+Audio) of target features varies greatly depending on the training source domains, increasing in the multi-source scenario (concept of ``unbalance", Section \ref{sec:intuition}).
However, despite the lack of a direct constraint across the sources, 
by 
minimizing RNA loss together with the classification loss, the network  
learns a set of weights (shared across all sources) that re-balances the contribution of the various input sources. In such a way, the network can exploit the information from all the input sources equally, as it has been demonstrated that norm mismatch between domains account for their erratic discrimination \cite{da-afnxu2019larger}.
We also noticed that by decreasing the total norm, the network promotes those features which are task-specific and meaningful to the final prediction, decreasing domain-specific ones which degrade performance on unseen data. This is shown in Figure \ref{fig:norm_unbalance}-b, where we plot the total norm of the Top-300\footnote{The Top-$300$ is obtained selecting the features corresponding to the 300-weights of the main classifier that mostly affect the final prediction.} features used for classification. By minimizing RNA loss, the percentage of this features have increased passing from up to 64\% to up to 72\% of the overall norm.

\textbf{Best Single-Source}. This experiment is a common practice in multi-source scenarios \cite{wang2020learning}. We choose the best source (the one with the highest performance) for each target, such as D2 for D3 (D2 $\rightarrow$ D3 $>$ D1 $\rightarrow$ D3) (Table \ref{tab:ablations}). With this experiment, we aim to show that i) as a multi-modal problem, having many sources do not necessarily guarantee an improvement (DeepAll $<$ Best Single Source), therefore the  need of using ad-hoc techniques to deal with multiple sources; ii) our loss allows the network to gain greater advantage from many sources (RNA-Net $>$ Best Single Source + RNA $>$ Best Single Source), confirming the domain generalization abilities of RNA-Net and the fact that it is not limited to tackle a multi-modal problem.


\noindent

\begin{table}[t!]
\centering

\begin{adjustbox}{width=\columnwidth, margin=0ex 1ex 0ex 0ex}
\begin{tabular}{l|cccc}

\toprule\noalign{\smallskip}
\multicolumn{5}{c}{\normalsize\textsc{Ablation Study}} \\
\noalign{\smallskip}
\cline{1-5}
\noalign{\smallskip}

    & \multicolumn{1}{c}{ D2, D3  $\rightarrow$ D1} &    D1, D3  $\rightarrow$ D2 &        D1, D2  $\rightarrow$ D3& Mean \\
 \noalign{\smallskip} \hline
 \noalign{\smallskip}

DeepAll  &      39.35   &   43.19   & 51.47         & 44.67           \\

HNA       & 49.41 & 45.88  &   50.24             & 48.51 \\
RNA$^{sub}$      &   49.97   & 47.88  & 53.33                & 50.39 \\

RNA      &  51.64 &   {45.65}    & 55.88   &                \textbf{51.06} \\ \hline
\rowcolor{Gray}
Supervised      &
\multicolumn{1}{c}{50.11}	&	\multicolumn{1}{c}{63.62}	&	\multicolumn{1}{c}{65.56}        	&	{59.76} \\
\rowcolor{Gray}
Supervised  +RNA    &  \multicolumn{1}{c}{54.68}	&	\multicolumn{1}{c}{67.48}	&	\multicolumn{1}{c}{67.24}                     	&	\textbf{63.13}
 \\
\bottomrule

\end{tabular}
\end{adjustbox}
\caption{Accuracy (\%) of HNA, RNA\textsuperscript{sub} and RNA losses proposed in the main paper.
In \textbf{bold} we show the highest mean result.}
\vspace{-0.5cm}
\label{tab:ablation_}
\end{table}

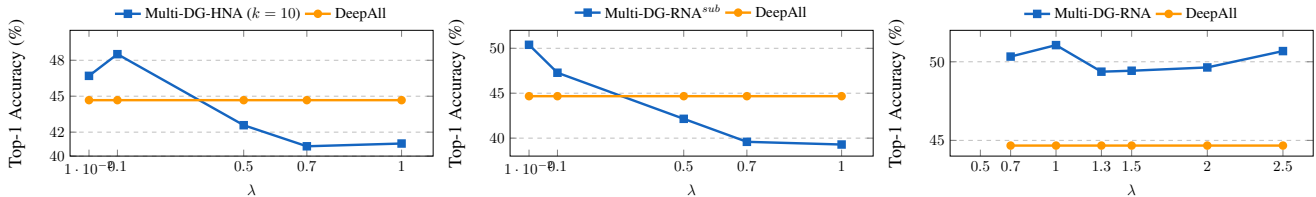
\begin{figure*}[t!]
\centering
    \begin{minipage}[t]{0.33\textwidth}
     \resizebox {\columnwidth} {!} {
        \begin{tikzpicture}
        \pgfplotsset{every axis legend/.append style={
at={(0.5,1.03)},
anchor=south}}

        \begin{axis}[
          enlargelimits=false,
          ylabel={\large Top-1 Accuracy (\%)},
          xlabel={ $\lambda$},
           xmin=-0.05, xmax=1.1,
           ymin=40, ymax=50.5,
          xtick={0.01,0.1,0.5,0.7, 1, 1.3},
          ytick={40,42,45,48,60,65},
          ymajorgrids=true,
          grid style=dashed,
          width=10cm,
          height=4.5cm,
          legend columns=-1,
        legend style={draw=none},
        every axis plot/.append style={ultra thick},
        every mark/.append style={mark size=50pt}
        ]
        \addplot[
          color=MaterialBlue800,
          mark=square*]
        table[x index=0,y index=1,col sep=comma]
        {latex/Tables/exp/multi-dg_hna.txt};
        \addplot[
          color=MaterialOrange500,
          mark=oplus*]
        table[x index=0,y index=1,col sep=comma]
        {latex/Tables/exp/deep-all-2.txt};
        \legend{Multi-DG-HNA ($k=10$), DeepAll} 
        
        \end{axis}

        \end{tikzpicture}%
      }

    \end{minipage}
    \begin{minipage}[t]{0.33\textwidth}
     \resizebox {\columnwidth} {!} {
        \begin{tikzpicture}
        \pgfplotsset{every axis legend/.append style={
at={(0.5,1.03)},
anchor=south}}
        \begin{axis}[
          enlargelimits=false,
          ylabel={\large Top-1 Accuracy (\%)},
          xlabel={$\lambda$},
           xmin=-0.05, xmax=1.1,
           ymin=38, ymax=52,
          xtick={0.01,0.1,0.5,0.7, 1},
          ytick={40,45,50,60,65},
          ymajorgrids=true,
          grid style=dashed,
          width=10cm,
          height=4.5cm,
          legend columns=-1,
           legend style={draw=none},
           every axis plot/.append style={ultra thick}
        ]
        \addplot[
          color=MaterialBlue800,
          mark=square*]
        table[x index=0,y index=1,col sep=comma]
        {latex/Tables/exp/multi-dg_sub.txt};
        \addplot[
          color=MaterialOrange500,
          mark=oplus*]
        table[x index=0,y index=1,col sep=comma]
        {latex/Tables/exp/deep-all-2.txt};
        \legend{Multi-DG-RNA$^{sub}$,DeepAll}

        \end{axis}

        \end{tikzpicture}%
      }
    \end{minipage}
    \begin{minipage}[t]{0.33\textwidth}
     \resizebox {\columnwidth} {!} {
        \begin{tikzpicture}
        \pgfplotsset{every axis legend/.append style={
at={(0.5,1.03)},
anchor=south}}
        \begin{axis}[
          enlargelimits=false,
          ylabel={\large Top-1 Accuracy (\%)},
          xlabel={$\lambda$},
           xmin=0.3, xmax=2.7,
           ymin=44, ymax=52,
          xtick={0.5,0.7,1,1.3,1.5, 2, 2.5},
          ytick={40,45,50,60,65},
          ymajorgrids=true,
          grid style=dashed,
          width=10cm,
          height=4.5cm,
          legend columns=-1,
           legend style={draw=none},
           every axis plot/.append style={ultra thick}
        ]
        \addplot[
          color=MaterialBlue800,
          mark=square*]
        table[x index=0,y index=1,col sep=comma]
        {latex/Tables/exp/multi-dg.txt};
        \addplot[
          color=MaterialOrange500,
          mark=oplus*]
        table[x index=0,y index=1,col sep=comma]
        {latex/Tables/exp/deep-all.txt};
        \legend{Multi-DG-RNA,DeepAll}

        \end{axis}

        \end{tikzpicture}%
      }
    \end{minipage}
    
\caption{Different performance (average Top-1 Accuracy (\%)) based on the value of $\lambda$ used to weight HNA and RNA$^{sub}$ and RNA losses.}
\label{fig:lambda_var_1}
\vspace{-0.4cm}
\end{figure*}


\textbf{Multi-Modal Approaches.} 
In Table \ref{tab:mm-1} we compare RNA-Net against recent audio-visual methods, which we adapted to our setting. 
This is to verify if increasing cooperation between the two modalities, through other strategies, still improves the network's generalization abilities.
Those are TBN \cite{Kazakos_2019_ICCV}, based on temporal aggregation, and two multi-modal Transformer-based approaches \cite{cheng2020look,morgado2020learning}.
Finally, since gating fusion approaches have been demonstrated to be valid multi-modal fusion strategies \cite{kiela2018efficient}, we adapted Squeeze And Excitation \cite{squeeze_and_excitation} and Non-Local \cite{non-local}. We leave details about the implementation in the Supplementary. These experiments confirm that by enhancing the audio-visual modality's cooperation the network's generalization improves. RNA-Net, on the other hand, surpasses all of those approaches by a large margin, demonstrating yet again how useful it is in cross-domain scenarios.

\textbf{DA Results.}  
Results in UDA, when target data (unlabeled) is available at training time, are summarized in Table \ref{tab:mm-2}. We validate RNA-Net against four existing UDA approaches, namely AdaBN \cite{ada-bn}, MMD \cite{da-mmdlong2015learning}, GRL \cite{grl-pmlr-v37-ganin15} and 
MM-SADA\footnote{To put MM-SADA \cite{Munro_2020_CVPR} on equal footing to RNA-Net, we run it with audio-visual input} \cite{Munro_2020_CVPR}.
 The baseline is the \textit{Source-Only} (training on source and testing directly on target data). MM-SADA \cite{Munro_2020_CVPR} combines a self-supervised approach (SS) with an adversarial one (GRL). We compare RNA-Net with both the complete method and its DA single components (SS, GRL). Interestingly, it provides comparable results to MM-SADA despite not being expressly designed as a UDA-based technique.  It should also be noted that MM-SADA must be trained in stages, while RNA-Net is end-to-end trainable. Finally, we prove the complementarity of our approach with the adversarial one (RNA-Net+GRL), achieving a slight improvement over MM-SADA.


\setlength{\tabcolsep}{2pt}
\begin{table}
\begin{minipage}[t]{0.46\columnwidth}
\begin{adjustbox}{width=\linewidth}
\begin{tabular}{lcccc}
\toprule
Method & S-DG & M-DG & Norm & Angle\\
 \noalign{\smallskip} \hline \noalign{\smallskip}
 DeepAll & 41.87 & 44.67 & &\\       
\hline
 \noalign{\smallskip}

CosSim & 41.76 & 45.60 & & \cmark  \\
MSE & 45.52 & 46.11   & \cmark   & \cmark   \\ 
Orth.Loss & 42.67 & 47.64   &&\cmark     \\ 
 \noalign{\smallskip}

\hline \noalign{\smallskip}

RNA loss & \textbf{45.75} & \textbf{51.06} & \cmark  &     \\ 
\bottomrule
\end{tabular}
\end{adjustbox}
\vspace{+0.05cm}
\caption{Comparison in terms of accuracy (\%) between RNA loss and other existing losses.}
\vspace{-1cm}

\label{tab:other_losses}
\end{minipage}
\hfill
\begin{minipage}[t]{0.49\columnwidth}
\begin{adjustbox}{width=\linewidth}
\begin{tabular}{l|ccc}
\toprule
\multicolumn{4}{c}{\normalsize\textsc{EpicKitchen-100}} \\
\noalign{\smallskip}
\cline{1-4}
\noalign{\smallskip}
  & \multicolumn{1}{c}{Target} &  Top-1 &  Top-5 \\ 
 \noalign{\smallskip} \hline \noalign{\smallskip}

Source Only & \xmark               & 44.39 & 69.69 \\ \hline \noalign{\smallskip}

TA$^3$N  \cite{videoda-chen2019temporal}      & \cmark               & 46.91 & 72.70 \\ \hline \noalign{\smallskip}

RNA-Net         & \xmark              & \underline{47.96} & \underline{79.54} \\ \hline
\noalign{\smallskip}
TA$^3$N \cite{videoda-chen2019temporal}+RNA-Net    & \cmark               & \textbf{50.40} & \textbf{80.47} \\ 
\bottomrule
\end{tabular}
\end{adjustbox}
\vspace{+0.05cm}
\caption{Verb accuracy (\%) on the EK-100 UDA setting.}

\label{tab:EK}
\end{minipage}
\end{table}

\textbf{Ablation Study.} \label{par:baselinemethods} 
In Table \ref{tab:ablations} 
we show the performance of the two modalities when trained separately (\textit{RGB Only}, \textit{Audio Only}) and tested directly on unseen data, showing that the fusion of the two streams provides better results. In Table \ref{tab:ablation_} we also perform an ablation study to validate the choices on the formulation of RNA loss. In particular, we compare it against the \textit{Hard Norm Alignment} loss (HNA), and against RNA$^{sub}$, confirming that the dividend/divisor structure is the one achieving better performance. Finally, we show that our loss not only benefits across domains, but also improves performance in the supervised setting.

\textbf{Ablation on $\lambda$ variations.}
In Figure \ref{fig:lambda_var_1}, we compare the performances of HNA, {RNA}$^{sub}$, and RNA respectively as a function of $\lambda$. Results show that the performance of both HNA and RNA$^{sub}$ are highly sensitive to $\lambda$. Specifically, higher values of $\lambda$ cause significant performance degradations since (potential) large difference values between $\EX[h(X^a)]$ and $\EX[h(X^v)]$ (for RNA$^{sub}$) or $k$ and $\EX[h(X^m)]$ (for HNA) result in high loss values that could cause the network to diverge. These convergence problems are softened by the ``ratio" structure of RNA, which outperforms the baseline results on all choices of $\lambda$. 

\textbf{Comparison with other losses.} 
We compare the RNA loss against a standard cosine similarity-based loss and an Euclidean-based loss, i.e., the Mean Square Error (MSE) (Table \ref{tab:other_losses}). The first enforces alignment by minimizing the angular distance between the two feature representations, and the second tends to align both their angular and norms by minimizing the $L2$-loss of the two. The results suggest that re-balancing the norms has a greater impact than not using angular limitations.
In fact, RNA (norm re-balance, no angular constraint) outperforms MSE (both norm re-balance and angular constraint), notably in multi-DG. 
We believe that a loss should allow feature distributions to retain modality-specific features when one modality is weak or contains only domain-information, and to align them when both are connected with action. To this purpose, we compare RNA loss to an orthogonality loss, which keeps modality-specific properties rather than aligning them (details on Supplementary). As shown in Table \ref{tab:other_losses}, the Orth. loss outperforms the CosSim (alignment) loss, especially in multi-DG, proving that utilizing modality-specific features 
deals with domain shift better. The RNA loss outperforms it by a significant margin. Our intuition is that by not constraining the angle, 
we do not strictly enforce an alignment (when non-necessary) or an orthogonality, letting the network to find itself the better angle for the main task.

\begin{figure}[t]
    \includegraphics[width=1\linewidth]{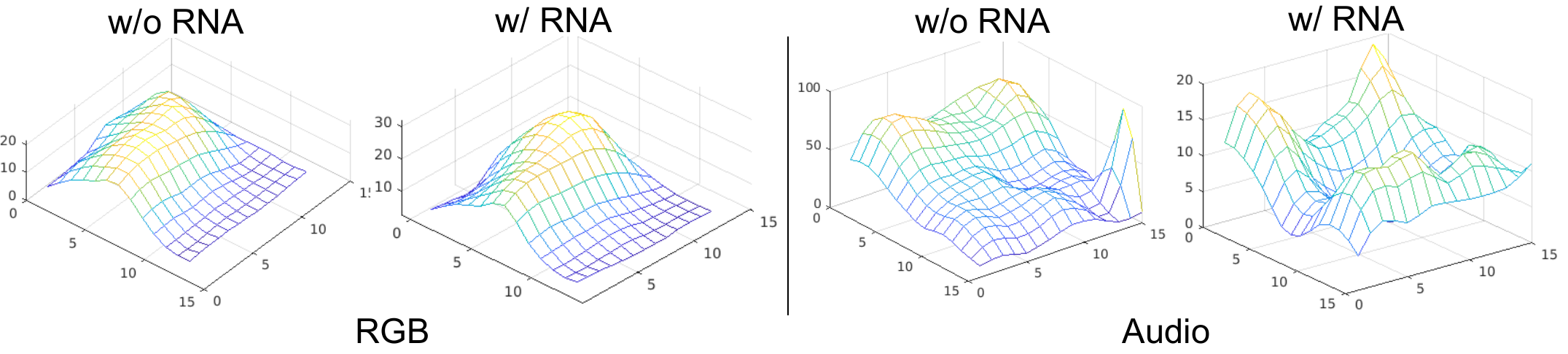}
    \caption{Norm ranges of RGB and audio. When minimizing the RNA loss, 
    relevant features are kept high, while the less relevant are decreased (see yellow peaks). Best viewed in colors.
} \vspace{-0.4cm}
    \label{fig:qualitative}
\end{figure}

\textbf{EPIC-Kitchens-100 UDA.} The results achieved on the recently proposed EPIC-Kitchens-100 UDA setting ~\cite{damen2020rescaling} are shown in Table \ref{tab:EK}. The source is composed of videos from EK-55, while the target is made up of videos from EK-100. 
RNA-Net outperforms the baseline Source Only by up to $3\%$ on Top-1 and $10\%$ on Top-5, remarking the importance of using ad-hoc  techniques to deal with multiple sources (see Supplementary). Moreover, it outperforms the very recent UDA technique TA$^3$N \cite{videoda-chen2019temporal} \textit{\textbf{without access to target data}}. 
Interestingly, when combined to TA$^3$N, it further improves performance, proving the complementarity of RNA-Net to other existing UDA approaches.

\textbf{Qualitative Analysis.} In Figure \ref{fig:qualitative} we empirically show the effect of RNA loss on feature norms, by analyzing the behaviour of the spatial features' norms, i.e., features before the last average pooling. When increasing the mean feature norms (in the case of RGB), the most significant features are increased, while when decreasing them (in the case of audio) the unrelevant (domain-specific) ones are reduced. This is evident especially in the case of audio (right). 

\label{tab:ablation}

\section{Conclusion}

In this paper we showed for the first time that generalization to unseen domains in first person action recognition can be achieved effectively by leveraging over the complementary nature of audio and visual modalities, bringing to light the ``norm unbalance" problem.
We presented an innovative vision on multi-modal research, proposing the modality feature norms as a measure unit. To this end, we designed a new cross-modal loss that operates directly on the relative feature norm of the two modalities.
We see this norm-based approach as a promising new take on multi-modal learning, potentially of interest for many other
research fields. 

{\small 
\textbf{Acknowledgements. }
The work was partially supported by the ERC project N. 637076 RoboExNovo and the research herein was carried out using the IIT HPC infrastructure. This work was supported by the CINI Consortium through the VIDESEC project.
}

\typeout{}
{\small
\bibliographystyle{ieee_fullname}
\bibliography{egbib}

\begin{thebibliography}{10}\itemsep=-1pt

\bibitem{afourasself}
Triantafyllos Afouras, Andrew Owens, Joon~Son Chung, and Andrew Zisserman.
\newblock Self-supervised learning of audio-visual objects from video.
\newblock In {\em Computer Vision--ECCV 2020: 16th European Conference,
  Glasgow, UK, August 23--28, 2020, Proceedings, Part XVIII 16}, pages
  208--224. Springer, 2020.

\bibitem{agarwal2020unsupervised}
Nakul Agarwal, Yi-Ting Chen, Behzad Dariush, and Ming-Hsuan Yang.
\newblock Unsupervised domain adaptation for spatio-temporal action
  localization.
\newblock {\em arXiv preprint arXiv:2010.09211}, 2020.

\bibitem{alamri2019audio}
Huda Alamri, Vincent Cartillier, Abhishek Das, Jue Wang, Anoop Cherian, Irfan
  Essa, Dhruv Batra, Tim~K Marks, Chiori Hori, Peter Anderson, et~al.
\newblock Audio visual scene-aware dialog.
\newblock In {\em Proceedings of the IEEE/CVF Conference on Computer Vision and
  Pattern Recognition}, pages 7558--7567, 2019.

\bibitem{alwassel2019self}
Humam Alwassel, Dhruv Mahajan, Bruno Korbar, Lorenzo Torresani, Bernard Ghanem,
  and Du Tran.
\newblock Self-supervised learning by cross-modal audio-video clustering.
\newblock {\em arXiv preprint arXiv:1911.12667}, 2019.

\bibitem{look_listen_learn}
Relja Arandjelovic and Andrew Zisserman.
\newblock Look, listen and learn.
\newblock In {\em Proceedings of the IEEE International Conference on Computer
  Vision}, pages 609--617, 2017.

\bibitem{objects_that_sound}
Relja Arandjelovic and Andrew Zisserman.
\newblock Objects that sound.
\newblock In {\em Proceedings of the European Conference on Computer Vision
  (ECCV)}, pages 435--451, 2018.

\bibitem{bucci2020selfsupervised}
Silvia Bucci, Antonio D'Innocente, Yujun Liao, Fabio~Maria Carlucci, Barbara
  Caputo, and Tatiana Tommasi.
\newblock Self-supervised learning across domains.
\newblock {\em IEEE Transactions on Pattern Analysis and Machine Intelligence},
  2021.

\bibitem{carlucci2019domain}
Fabio~M Carlucci, Antonio D'Innocente, Silvia Bucci, Barbara Caputo, and
  Tatiana Tommasi.
\newblock Domain generalization by solving jigsaw puzzles.
\newblock In {\em Proceedings of the IEEE Conference on Computer Vision and
  Pattern Recognition}, pages 2229--2238, 2019.

\bibitem{carreira2017quo}
Joao Carreira and Andrew Zisserman.
\newblock Quo vadis, action recognition? a new model and the kinetics dataset.
\newblock In {\em proceedings of the IEEE Conference on Computer Vision and
  Pattern Recognition}, pages 6299--6308, 2017.

\bibitem{cartas2019seeing}
Alejandro Cartas, Jordi Luque, Petia Radeva, Carlos Segura, and Mariella
  Dimiccoli.
\newblock Seeing and hearing egocentric actions: How much can we learn?
\newblock In {\em Proceedings of the IEEE International Conference on Computer
  Vision Workshops}, pages 0--0, 2019.

\bibitem{da-bnchang2019domain}
Woong-Gi Chang, Tackgeun You, Seonguk Seo, Suha Kwak, and Bohyung Han.
\newblock Domain-specific batch normalization for unsupervised domain
  adaptation.
\newblock In {\em Proceedings of the IEEE Conference on Computer Vision and
  Pattern Recognition}, pages 7354--7362, 2019.

\bibitem{videoda-chen2019temporal}
Min-Hung Chen, Zsolt Kira, Ghassan AlRegib, Jaekwon Yoo, Ruxin Chen, and Jian
  Zheng.
\newblock Temporal attentive alignment for large-scale video domain adaptation.
\newblock In {\em Proceedings of the IEEE International Conference on Computer
  Vision}, pages 6321--6330, 2019.

\bibitem{chen2020action}
Min-Hung Chen, Baopu Li, Yingze Bao, Ghassan AlRegib, and Zsolt Kira.
\newblock Action segmentation with joint self-supervised temporal domain
  adaptation.
\newblock In {\em Proceedings of the IEEE/CVF Conference on Computer Vision and
  Pattern Recognition}, pages 9454--9463, 2020.

\bibitem{cheng2020look}
Ying Cheng, Ruize Wang, Zhihao Pan, Rui Feng, and Yuejie Zhang.
\newblock Look, listen, and attend: Co-attention network for self-supervised
  audio-visual representation learning.
\newblock In {\em Proceedings of the 28th ACM International Conference on
  Multimedia}, pages 3884--3892, 2020.

\bibitem{videoda-choi2020unsupervised}
Jinwoo Choi, Gaurav Sharma, Manmohan Chandraker, and Jia-Bin Huang.
\newblock Unsupervised and semi-supervised domain adaptation for action
  recognition from drones.
\newblock In {\em The IEEE Winter Conference on Applications of Computer
  Vision}, pages 1717--1726, 2020.

\bibitem{chung2016out}
Joon~Son Chung and Andrew Zisserman.
\newblock Out of time: automated lip sync in the wild.
\newblock In {\em Asian conference on computer vision}, pages 251--263.
  Springer, 2016.

\bibitem{Crasto_2019_CVPR}
Nieves Crasto, Philippe Weinzaepfel, Karteek Alahari, and Cordelia Schmid.
\newblock Mars: Motion-augmented rgb stream for action recognition.
\newblock In {\em Proceedings of the IEEE/CVF Conference on Computer Vision and
  Pattern Recognition (CVPR)}, June 2019.

\bibitem{damen2018scaling}
Dima Damen, Hazel Doughty, Giovanni~Maria Farinella, Sanja Fidler, Antonino
  Furnari, Evangelos Kazakos, Davide Moltisanti, Jonathan Munro, Toby Perrett,
  Will Price, et~al.
\newblock Scaling egocentric vision: The epic-kitchens dataset.
\newblock In {\em Proceedings of the European Conference on Computer Vision
  (ECCV)}, pages 720--736, 2018.

\bibitem{damen2020rescaling}
Dima Damen, Hazel Doughty, Giovanni~Maria Farinella, Antonino Furnari,
  Evangelos Kazakos, Jian Ma, Davide Moltisanti, Jonathan Munro, Toby Perrett,
  Will Price, et~al.
\newblock Rescaling egocentric vision.
\newblock {\em arXiv preprint arXiv:2006.13256}, 2020.

\bibitem{ek2020}
Dima Damen, Evangelos Kazakos, Will Price, Jian Ma, and Hazel Doughty.
\newblock Epic-kitchens-55 - 2020 challenges report.
\newblock
  \url{https://epic-kitchens.github.io/Reports/EPIC-KITCHENS-Challenges-2020-Report.pdf},
  2020.

\bibitem{ek19report}
Dima Damen, Will Price, Evangelos Kazakos, Antonino Furnari, and Giovanni~Maria
  Farinella.
\newblock Epic-kitchens - 2019 challenges report.
\newblock
  \url{https://epic-kitchens.github.io/Reports/EPIC-Kitchens-Challenges-2019-Report.pdf},
  2019.

\bibitem{imageNet}
J. {Deng}, W. {Dong}, R. {Socher}, L. {Li}, {Kai Li}, and {Li Fei-Fei}.
\newblock Imagenet: A large-scale hierarchical image database.
\newblock In {\em 2009 IEEE Conference on Computer Vision and Pattern
  Recognition}, pages 248--255, 2009.

\bibitem{da-adv-deng2019cluster}
Zhijie Deng, Yucen Luo, and Jun Zhu.
\newblock Cluster alignment with a teacher for unsupervised domain adaptation.
\newblock In {\em Proceedings of the IEEE International Conference on Computer
  Vision}, pages 9944--9953, 2019.

\bibitem{dou2019domain}
Qi Dou, Daniel Coelho~de Castro, Konstantinos Kamnitsas, and Ben Glocker.
\newblock Domain generalization via model-agnostic learning of semantic
  features.
\newblock {\em Advances in Neural Information Processing Systems},
  32:6450--6461, 2019.

\bibitem{fathi2011learning}
Alireza Fathi, Xiaofeng Ren, and James~M Rehg.
\newblock Learning to recognize objects in egocentric activities.
\newblock In {\em CVPR 2011}, pages 3281--3288. IEEE, 2011.

\bibitem{feichtenhofer2019slowfast}
Christoph Feichtenhofer, Haoqi Fan, Jitendra Malik, and Kaiming He.
\newblock Slowfast networks for video recognition.
\newblock In {\em Proceedings of the IEEE/CVF international conference on
  computer vision}, pages 6202--6211, 2019.

\bibitem{furnari2020rolling}
Antonino Furnari and Giovanni Farinella.
\newblock Rolling-unrolling lstms for action anticipation from first-person
  video.
\newblock {\em IEEE Transactions on Pattern Analysis and Machine Intelligence},
  2020.

\bibitem{grl-pmlr-v37-ganin15}
Yaroslav Ganin and Victor Lempitsky.
\newblock Unsupervised domain adaptation by backpropagation.
\newblock volume~37 of {\em Proceedings of Machine Learning Research}, pages
  1180--1189, Lille, France, 07--09 Jul 2015. PMLR.

\bibitem{listen_to_look}
Ruohan Gao, Tae-Hyun Oh, Kristen Grauman, and Lorenzo Torresani.
\newblock Listen to look: Action recognition by previewing audio.
\newblock In {\em Proceedings of the IEEE/CVF Conference on Computer Vision and
  Pattern Recognition}, pages 10457--10467, 2020.

\bibitem{ghadiyaram2019large}
Deepti Ghadiyaram, Du Tran, and Dhruv Mahajan.
\newblock Large-scale weakly-supervised pre-training for video action
  recognition.
\newblock In {\em Proceedings of the IEEE Conference on Computer Vision and
  Pattern Recognition}, pages 12046--12055, 2019.

\bibitem{goyal2017making}
Yash Goyal, Tejas Khot, Douglas Summers-Stay, Dhruv Batra, and Devi Parikh.
\newblock Making the v in vqa matter: Elevating the role of image understanding
  in visual question answering.
\newblock In {\em Proceedings of the IEEE Conference on Computer Vision and
  Pattern Recognition}, pages 6904--6913, 2017.

\bibitem{squeeze_and_excitation}
Jie Hu, Li Shen, and Gang Sun.
\newblock Squeeze-and-excitation networks.
\newblock In {\em Proceedings of the IEEE conference on computer vision and
  pattern recognition}, pages 7132--7141, 2018.

\bibitem{bn-inception}
Sergey Ioffe and Christian Szegedy.
\newblock Batch normalization: Accelerating deep network training by reducing
  internal covariate shift.
\newblock pages 448--456, 2015.

\bibitem{videoda-Jamal2018DeepDA}
A. Jamal, Vinay~P. Namboodiri, Dipti Deodhare, and K. Venkatesh.
\newblock Deep domain adaptation in action space.
\newblock In {\em BMVC}, 2018.

\bibitem{kapidis2019multitask}
Georgios Kapidis, Ronald Poppe, Elsbeth van Dam, Lucas Noldus, and Remco
  Veltkamp.
\newblock Multitask learning to improve egocentric action recognition.
\newblock In {\em Proceedings of the IEEE International Conference on Computer
  Vision Workshops}, pages 0--0, 2019.

\bibitem{Kazakos_2019_ICCV}
Evangelos Kazakos, Arsha Nagrani, Andrew Zisserman, and Dima Damen.
\newblock Epic-fusion: Audio-visual temporal binding for egocentric action
  recognition.
\newblock In {\em The IEEE International Conference on Computer Vision (ICCV)},
  October 2019.

\bibitem{kazakos2021slow}
Evangelos Kazakos, Arsha Nagrani, Andrew Zisserman, and Dima Damen.
\newblock Slow-fast auditory streams for audio recognition.
\newblock In {\em ICASSP 2021-2021 IEEE International Conference on Acoustics,
  Speech and Signal Processing (ICASSP)}, pages 855--859. IEEE, 2021.

\bibitem{kendall2018multi}
Alex Kendall, Yarin Gal, and Roberto Cipolla.
\newblock Multi-task learning using uncertainty to weigh losses for scene
  geometry and semantics.
\newblock In {\em Proceedings of the IEEE conference on computer vision and
  pattern recognition}, pages 7482--7491, 2018.

\bibitem{kiela2018efficient}
Douwe Kiela, Edouard Grave, Armand Joulin, and Tomas Mikolov.
\newblock Efficient large-scale multi-modal classification.
\newblock In {\em Proceedings of the AAAI Conference on Artificial
  Intelligence}, volume~32, 2018.

\bibitem{korbar2018co}
Bruno Korbar.
\newblock Co-training of audio and video representations from self-supervised
  temporal synchronization.
\newblock 2018.

\bibitem{cooperative_torresani}
Bruno Korbar, Du Tran, and Lorenzo Torresani.
\newblock Cooperative learning of audio and video models from self-supervised
  synchronization.
\newblock In {\em Proceedings of the 32nd International Conference on Neural
  Information Processing Systems}, NIPS'18, page 7774–7785. Curran Associates
  Inc., 2018.

\bibitem{korbar2018cooperative}
Bruno Korbar, Du Tran, and Lorenzo Torresani.
\newblock Cooperative learning of audio and video models from self-supervised
  synchronization.
\newblock {\em arXiv preprint arXiv:1807.00230}, 2018.

\bibitem{kumar2018wearable}
Nallapaneni~Manoj Kumar, Neeraj~Kumar Singh, and VK Peddiny.
\newblock Wearable smart glass: Features, applications, current progress and
  challenges.
\newblock In {\em 2018 Second International Conference on Green Computing and
  Internet of Things (ICGCIoT)}, pages 577--582. IEEE, 2018.

\bibitem{lee2018motion}
Myunggi Lee, Seungeui Lee, Sungjoon Son, Gyutae Park, and Nojun Kwak.
\newblock Motion feature network: Fixed motion filter for action recognition.
\newblock In {\em Proceedings of the European Conference on Computer Vision
  (ECCV)}, pages 387--403, 2018.

\bibitem{li2018domain}
Haoliang Li, Sinno Jialin~Pan, Shiqi Wang, and Alex~C Kot.
\newblock Domain generalization with adversarial feature learning.
\newblock In {\em Proceedings of the IEEE Conference on Computer Vision and
  Pattern Recognition}, pages 5400--5409, 2018.

\bibitem{li2018eye}
Yin Li, Miao Liu, and James~M Rehg.
\newblock In the eye of beholder: Joint learning of gaze and actions in first
  person video.
\newblock In {\em Proceedings of the European Conference on Computer Vision
  (ECCV)}, pages 619--635, 2018.

\bibitem{li2018deep}
Ya Li, Xinmei Tian, Mingming Gong, Yajing Liu, Tongliang Liu, Kun Zhang, and
  Dacheng Tao.
\newblock Deep domain generalization via conditional invariant adversarial
  networks.
\newblock In {\em Proceedings of the European Conference on Computer Vision
  (ECCV)}, pages 624--639, 2018.

\bibitem{ada-bn}
Yanghao Li, Naiyan Wang, Jianping Shi, Xiaodi Hou, and Jiaying Liu.
\newblock Adaptive batch normalization for practical domain adaptation.
\newblock {\em Pattern Recognition}, 80:109--117, 2018.

\bibitem{DBLP:conf/iclr/LiWS0H17}
Yanghao Li, Naiyan Wang, Jianping Shi, Jiaying Liu, and Xiaodi Hou.
\newblock Revisiting batch normalization for practical domain adaptation.
\newblock In {\em 5th International Conference on Learning Representations,
  {ICLR} 2017, Toulon, France, April 24-26, 2017, Workshop Track Proceedings}.
  OpenReview.net, 2017.

\bibitem{lin2019tsm}
Ji Lin, Chuang Gan, and Song Han.
\newblock Tsm: Temporal shift module for efficient video understanding.
\newblock In {\em Proceedings of the IEEE International Conference on Computer
  Vision}, pages 7083--7093, 2019.

\bibitem{da-mmdlong2015learning}
Mingsheng Long, Yue Cao, Jianmin Wang, and Michael Jordan.
\newblock Learning transferable features with deep adaptation networks.
\newblock In {\em International conference on machine learning}, pages 97--105.
  PMLR, 2015.

\bibitem{ma2016going}
Minghuang Ma, Haoqi Fan, and Kris~M Kitani.
\newblock Going deeper into first-person activity recognition.
\newblock In {\em Proceedings of the IEEE Conference on Computer Vision and
  Pattern Recognition}, pages 1894--1903, 2016.

\bibitem{morgado2020learning}
Pedro Morgado, Yi Li, and Nuno Vasconcelos.
\newblock Learning representations from audio-visual spatial alignment.
\newblock {\em arXiv preprint arXiv:2011.01819}, 2020.

\bibitem{Morgado_2021_CVPR}
Pedro Morgado, Ishan Misra, and Nuno Vasconcelos.
\newblock Robust audio-visual instance discrimination.
\newblock In {\em Proceedings of the IEEE/CVF Conference on Computer Vision and
  Pattern Recognition (CVPR)}, pages 12934--12945, June 2021.

\bibitem{morgado2021audio}
Pedro Morgado, Nuno Vasconcelos, and Ishan Misra.
\newblock Audio-visual instance discrimination with cross-modal agreement.
\newblock In {\em Proceedings of the IEEE/CVF Conference on Computer Vision and
  Pattern Recognition}, pages 12475--12486, 2021.

\bibitem{Munro_2020_CVPR}
Jonathan Munro and Dima Damen.
\newblock Multi-modal domain adaptation for fine-grained action recognition.
\newblock In {\em Proceedings of the IEEE/CVF Conference on Computer Vision and
  Pattern Recognition (CVPR)}, June 2020.

\bibitem{instance_bn}
Hyeonseob Nam and Hyo-Eun Kim.
\newblock Batch-instance normalization for adaptively style-invariant neural
  networks.
\newblock {\em arXiv preprint arXiv:1805.07925}, 2018.

\bibitem{multisensory_owens}
Andrew Owens and Alexei~A Efros.
\newblock Audio-visual scene analysis with self-supervised multisensory
  features.
\newblock In {\em Proceedings of the European Conference on Computer Vision
  (ECCV)}, pages 631--648, 2018.

\bibitem{pan2020adversarial}
Boxiao Pan, Zhangjie Cao, Ehsan Adeli, and Juan~Carlos Niebles.
\newblock Adversarial cross-domain action recognition with co-attention.
\newblock In {\em Proceedings of the AAAI Conference on Artificial
  Intelligence}, volume~34, pages 11815--11822, 2020.

\bibitem{planamente2021self}
Mirco Planamente, Andrea Bottino, and Barbara Caputo.
\newblock Self-supervised joint encoding of motion and appearance for first
  person action recognition.
\newblock In {\em 2020 25th International Conference on Pattern Recognition
  (ICPR)}, pages 8751--8758. IEEE, 2021.

\bibitem{poliak2018hypothesis}
Adam Poliak, Jason Naradowsky, Aparajita Haldar, Rachel Rudinger, and Benjamin
  Van~Durme.
\newblock Hypothesis only baselines in natural language inference.
\newblock {\em arXiv preprint arXiv:1805.01042}, 2018.

\bibitem{ranjan2017l2}
Rajeev Ranjan, Carlos~D Castillo, and Rama Chellappa.
\newblock L2-constrained softmax loss for discriminative face verification.
\newblock {\em arXiv preprint arXiv:1703.09507}, 2017.

\bibitem{rodin2021predicting}
Ivan Rodin, Antonino Furnari, Dimitrios Mavroedis, and Giovanni~Maria
  Farinella.
\newblock Predicting the future from first person (egocentric) vision: A
  survey.
\newblock {\em Computer Vision and Image Understanding}, page 103252, 2021.

\bibitem{da-mcdsaito2018maximum}
Kuniaki Saito, Kohei Watanabe, Yoshitaka Ushiku, and Tatsuya Harada.
\newblock Maximum classifier discrepancy for unsupervised domain adaptation.
\newblock In {\em Proceedings of the IEEE Conference on Computer Vision and
  Pattern Recognition}, pages 3723--3732, 2018.

\bibitem{senocak2018learning}
Arda Senocak, Tae-Hyun Oh, Junsik Kim, Ming-Hsuan Yang, and In~So Kweon.
\newblock Learning to localize sound source in visual scenes.
\newblock In {\em Proceedings of the IEEE Conference on Computer Vision and
  Pattern Recognition}, pages 4358--4366, 2018.

\bibitem{10.5555/2968826.2968890}
Karen Simonyan and Andrew Zisserman.
\newblock Two-stream convolutional networks for action recognition in videos.
\newblock In {\em Proceedings of the 27th International Conference on Neural
  Information Processing Systems - Volume 1}, NIPS'14, page 568–576,
  Cambridge, MA, USA, 2014. MIT Press.

\bibitem{singh2016first}
Suriya Singh, Chetan Arora, and CV Jawahar.
\newblock First person action recognition using deep learned descriptors.
\newblock In {\em Proceedings of the IEEE Conference on Computer Vision and
  Pattern Recognition}, pages 2620--2628, 2016.

\bibitem{Song_2021_CVPR}
Xiaolin Song, Sicheng Zhao, Jingyu Yang, Huanjing Yue, Pengfei Xu, Runbo Hu,
  and Hua Chai.
\newblock Spatio-temporal contrastive domain adaptation for action recognition.
\newblock In {\em Proceedings of the IEEE/CVF Conference on Computer Vision and
  Pattern Recognition (CVPR)}, pages 9787--9795, June 2021.

\bibitem{sudhakaran2019lsta}
Swathikiran Sudhakaran, Sergio Escalera, and Oswald Lanz.
\newblock Lsta: Long short-term attention for egocentric action recognition.
\newblock In {\em Proceedings of the IEEE Conference on Computer Vision and
  Pattern Recognition}, pages 9954--9963, 2019.

\bibitem{sudhakaran2020gate}
Swathikiran Sudhakaran, Sergio Escalera, and Oswald Lanz.
\newblock Gate-shift networks for video action recognition.
\newblock In {\em Proceedings of the IEEE/CVF Conference on Computer Vision and
  Pattern Recognition}, pages 1102--1111, 2020.

\bibitem{Sudhakaran_2017_ICCV}
Swathikiran Sudhakaran and Oswald Lanz.
\newblock Convolutional long short-term memory networks for recognizing first
  person interactions.
\newblock In {\em Proceedings of the IEEE International Conference on Computer
  Vision (ICCV) Workshops}, Oct 2017.

\bibitem{sudhakaran2018attention}
Swathikiran Sudhakaran and Oswald Lanz.
\newblock Attention is all we need: Nailing down object-centric attention for
  egocentric activity recognition.
\newblock {\em arXiv preprint arXiv:1807.11794}, 2018.

\bibitem{sun2018optical}
Shuyang Sun, Zhanghui Kuang, Lu Sheng, Wanli Ouyang, and Wei Zhang.
\newblock Optical flow guided feature: A fast and robust motion representation
  for video action recognition.
\newblock In {\em Proceedings of the IEEE conference on computer vision and
  pattern recognition}, pages 1390--1399, 2018.

\bibitem{da-adv-tang2020discriminative}
Hui Tang and Kui Jia.
\newblock Discriminative adversarial domain adaptation.
\newblock In {\em AAAI}, pages 5940--5947, 2020.

\bibitem{tian2020unified}
Yapeng Tian, Dingzeyu Li, and Chenliang Xu.
\newblock Unified multisensory perception: weakly-supervised audio-visual video
  parsing.
\newblock {\em arXiv preprint arXiv:2007.10558}, 2020.

\bibitem{tian2018audio}
Yapeng Tian, Jing Shi, Bochen Li, Zhiyao Duan, and Chenliang Xu.
\newblock Audio-visual event localization in unconstrained videos.
\newblock In {\em Proceedings of the European Conference on Computer Vision
  (ECCV)}, pages 247--263, 2018.

\bibitem{torralba2011unbiased}
Antonio Torralba and Alexei~A Efros.
\newblock Unbiased look at dataset bias.
\newblock In {\em CVPR 2011}, pages 1521--1528. IEEE, 2011.

\bibitem{tran2015learning}
Du Tran, Lubomir Bourdev, Rob Fergus, Lorenzo Torresani, and Manohar Paluri.
\newblock Learning spatiotemporal features with 3d convolutional networks.
\newblock In {\em Proceedings of the IEEE international conference on computer
  vision}, pages 4489--4497, 2015.

\bibitem{volpi2018generalizing}
Riccardo Volpi, Hongseok Namkoong, Ozan Sener, John~C Duchi, Vittorio Murino,
  and Silvio Savarese.
\newblock Generalizing to unseen domains via adversarial data augmentation.
\newblock In {\em Advances in neural information processing systems}, pages
  5334--5344, 2018.

\bibitem{wang2017normface}
Feng Wang, Xiang Xiang, Jian Cheng, and Alan~Loddon Yuille.
\newblock Normface: L2 hypersphere embedding for face verification.
\newblock In {\em Proceedings of the 25th ACM international conference on
  Multimedia}, pages 1041--1049, 2017.

\bibitem{wang2020learning}
Hang Wang, Minghao Xu, Bingbing Ni, and Wenjun Zhang.
\newblock Learning to combine: Knowledge aggregation for multi-source domain
  adaptation.
\newblock In {\em European Conference on Computer Vision}, pages 727--744.
  Springer, 2020.

\bibitem{wang2016temporal}
Limin Wang, Yuanjun Xiong, Zhe Wang, Yu Qiao, Dahua Lin, Xiaoou Tang, and Luc
  Van~Gool.
\newblock Temporal segment networks: Towards good practices for deep action
  recognition.
\newblock In {\em European conference on computer vision}, pages 20--36.
  Springer, 2016.

\bibitem{gradient-blending}
Weiyao Wang, Du Tran, and Matt Feiszli.
\newblock What makes training multi-modal classification networks hard?
\newblock In {\em Proceedings of the IEEE/CVF Conference on Computer Vision and
  Pattern Recognition}, pages 12695--12705, 2020.

\bibitem{non-local}
Xiaolong Wang, Ross Girshick, Abhinav Gupta, and Kaiming He.
\newblock Non-local neural networks.
\newblock In {\em Proceedings of the IEEE conference on computer vision and
  pattern recognition}, pages 7794--7803, 2018.

\bibitem{weston2011wsabie}
Jason Weston, Samy Bengio, and Nicolas Usunier.
\newblock Wsabie: Scaling up to large vocabulary image annotation.
\newblock In {\em Twenty-Second International Joint Conference on Artificial
  Intelligence}, 2011.

\bibitem{Wu_2019_CVPR}
Chao-Yuan Wu, Christoph Feichtenhofer, Haoqi Fan, Kaiming He, Philipp
  Krahenbuhl, and Ross Girshick.
\newblock Long-term feature banks for detailed video understanding.
\newblock In {\em Proceedings of the IEEE/CVF Conference on Computer Vision and
  Pattern Recognition (CVPR)}, June 2019.

\bibitem{da-afnxu2019larger}
Ruijia Xu, Guanbin Li, Jihan Yang, and Liang Lin.
\newblock Larger norm more transferable: An adaptive feature norm approach for
  unsupervised domain adaptation.
\newblock In {\em Proceedings of the IEEE International Conference on Computer
  Vision}, pages 1426--1435, 2019.

\bibitem{videodg-yao2019adversarial}
Zhiyu Yao, Yunbo Wang, Xingqiang Du, Mingsheng Long, and Jianmin Wang.
\newblock Adversarial pyramid network for video domain generalization.
\newblock {\em arXiv preprint arXiv:1912.03716}, 2019.

\bibitem{ye2018rethinking}
Jianbo Ye, Xin Lu, Zhe Lin, and James~Z Wang.
\newblock Rethinking the smaller-norm-less-informative assumption in channel
  pruning of convolution layers.
\newblock {\em ICLR}, 2018.

\bibitem{Zhao_2018_ECCV}
Hang Zhao, Chuang Gan, Andrew Rouditchenko, Carl Vondrick, Josh McDermott, and
  Antonio Torralba.
\newblock The sound of pixels.
\newblock In {\em Proceedings of the European Conference on Computer Vision
  (ECCV)}, September 2018.

\bibitem{zhao2019dance}
Jiaojiao Zhao and Cees~GM Snoek.
\newblock Dance with flow: Two-in-one stream action detection.
\newblock In {\em Proceedings of the IEEE Conference on Computer Vision and
  Pattern Recognition}, pages 9935--9944, 2019.

\bibitem{zheng2018ring}
Yutong Zheng, Dipan~K Pal, and Marios Savvides.
\newblock Ring loss: Convex feature normalization for face recognition.
\newblock In {\em Proceedings of the IEEE conference on computer vision and
  pattern recognition}, pages 5089--5097, 2018.

\bibitem{zhou2018temporal}
Bolei Zhou, Alex Andonian, Aude Oliva, and Antonio Torralba.
\newblock Temporal relational reasoning in videos.
\newblock In {\em Proceedings of the European Conference on Computer Vision
  (ECCV)}, pages 803--818, 2018.

\end{thebibliography}
}

\end{document}